\documentclass[submission,copyright,creativecommons]{eptcs}
\providecommand{\event}{CAPNS 2018} % Name of the event you are submitting to
\usepackage{breakurl}             % Not needed if you use pdflatex only.
\usepackage{graphicx}
\usepackage{cancel}
\usepackage{mathptmx}      % use Times fonts if available on your TeX system
\usepackage{proof}
\usepackage{amssymb}
\usepackage{amsmath}  
\usepackage{stmaryrd}
\usepackage{extarrows}
\usepackage{tikz}
\usepackage{array}
\usetikzlibrary{positioning}
\usetikzlibrary{matrix}
\usepackage{lscape}
\usepackage{booktabs}
\newcommand{\ra}[1]{\renewcommand{\arraystretch}{#1}}

\newcommand{\quotes}[1]{``#1''}
\newcommand{\singlequotes}[1]{`#1'}
\newcommand{\pijl}{\rightarrow}
\newcommand{\comp}{\circ}
\newcommand{\resright}{\triangleright}
\newcommand{\resleft}{\triangleleft}
\newcommand{\resdia}{\triangledown}
\newcommand{\resdiainv}{\resdia^{-1}}
\newcommand{\resrightinv}{\resright^{-1}}
\newcommand{\resleftinv}{\resleft^{-1}}
\newcommand{\tensor}{\otimes}
\newcommand{\bs}{\backslash}
\newcommand{\s}{\slash}
\newcommand{\fdia}{\Diamond}
\newcommand{\gbox}{\Box}
\newcommand{\arr}[3]{#1:#2\longrightarrow #3}

\newcommand{\Arrow}[1]{\xlongrightarrow{\displaystyle #1}}
\newcommand{\reals}{\mathbb{R}}

\newcommand{\Xlefta}{\widehat{\alpha}_{\diamond}^{l}}

\newcommand{\Xrighta}{\widehat{\alpha}_{\diamond}^{r}}

\newcommand{\F}[1]{\lceil #1 \rceil}
\newcommand{\G}[1]{\lfloor #1 \rfloor}
\newcommand{\fboxx}[1]{#1}
\newcommand{\redbox}[1]{\color{red}{#1}\color{black}{}}

\newcommand{\thickoverline}[1]{\overline{#1}} % TODO: Configure this to make the overline thicker

\newcommand{\KroneckerRho}[3]{\rho\kern-0.2em{\raisebox{0.6em}{ {\tiny $#1$} }} \kern-0.7em{\raisebox{0.15em}{ {\tiny $#2$} }} \kern-0.75em{\raisebox{-0.4em}{ {\tiny $#3$} }}}

\newcommand{\abs}[2]{\lambda #1. #2}
\newcommand{\leftapp}[2]{(#2 \ltimes #1)}
\newcommand{\rightapp}[2]{(#1 \rtimes #2)}

\newcommand{\lpair}[2]{\langle #1, #2\rangle}

\newcommand\deriveZero[1]{%
  \input{derivations/#1}%
}

\newcommand\deriveOne[2]{%
  \def\SUBPROOF{#2}%
  \input{derivations/#1}%
}

\newcommand{\formulaone}{(np \bs s) \s np \tensor (\fdia np \bs (np \s n) \tensor n)}
\newcommand{\formulatwo}{(np \bs s) \s np \tensor ((\fdia np \tensor \fdia np \bs (np / n)) \tensor n)}
\newcommand{\formulathree}{np}

\begin{document}

\title{Classical Copying versus Quantum Entanglement in Natural Language: The Case of VP-ellipsis}
%Frobenius Algebras in Syntax: an Account of Ellipsis in a Vector Semantics for Lambek Calculus with Limited Contraction

\def\titlerunning{Classical Copying Versus Quantum Entanglement in Natural Language}        % if too long for running head

\author{Gijs Wijnholds
	\institute{School of Electronic Engineering and Computer Science\\Queen Mary University of London
         \email{g.j.wijnholds@qmul.ac.uk}}
         \and
         Mehrnoosh Sadrzadeh
         \institute{School of Electronic Engineering and Computer Science\\Queen Mary University of London
         \email{mehrnoosh.sadrzadeh@qmul.ac.uk}}
}

\def\authorrunning{G.~J.~Wijnholds \& M.~Sadrzadeh} % if too long for running head

\date{Received: date / Accepted: date}

\maketitle

\begin{abstract} This paper compares classical copying and quantum entanglement in natural language by considering the case of verb phrase (VP) ellipsis. VP ellipsis is a non-linear linguistic phenomenon that requires the reuse of resources, making it the ideal test case for a comparative study of different copying behaviours in compositional models of natural language. Following the line of research in compositional distributional semantics set out by \cite{coecke2010mathematical} we develop an extension of the Lambek calculus which admits a controlled form of contraction to deal with the copying of linguistic resources. We then develop two different compositional models of distributional meaning for this calculus. In the first model,  we follow the categorical approach of \cite{coecke2013lambek} in which a functorial passage sends the proofs of the grammar to linear maps on vector spaces and we use Frobenius algebras to allow for copying. In the second case, we follow the more traditional approach that one finds in categorial grammars, whereby an intermediate step interprets proofs as non-linear lambda terms, using multiple variable occurrences that model classical copying. As a case study, we apply the models to  derive different readings of ambiguous elliptical phrases and compare the analyses that each model provides. 
%We subsequently  compare the two models by experimenting on a verb disambiguation dataset with elliptical phrase pairs.
%attempts to integrate typelogical accounts of pronoun relativisation and ellipsis in a distributional semantic setting. Distributional semantics has a lot to say about the statistical information of content words, but provides little guidance on how to treat function words. Formal semantics, on the other hand, has powerful mechanisms for dealing with relative pronouns, coordinators, and the like. We consider previous investigations into relative pronouns, coordination and ellipsis in the DisCoCat framework, and aim to unify them in a single theory of typelogical distributional semantics.
\end{abstract}

\section{Introduction}
\label{intro}

Lexical distributional models of meaning assume that the meaning of a word is given by its context, an idea that can be operationalised by deriving vectorial word representations from corpus co-occurrence statistics. These single-word embeddings by now are an essential part of the computational linguistics toolkit, as they have been successfully applied in several NLP tasks (see e.g. \cite{turney2010frequency, clark2015vector} for an overview). However, the move from single-word embeddings to larger phrases and full sentences poses a number of immediate issues. Firstly, the distributional hypothesis underlying the single-word embeddings does not  directly   apply to sentences: a sentence meaning is given by more than just its contexts. Second, such an approach would suffer from  data sparsity: although there has been  investigation in deriving vector representations for adjective-noun, verb-object, and  subject-verb-object  combinations from a corpus, e.g. see  \cite{lin2001dirt}, current corpora are not able to provide enough examples to successfully represent phrases.
%Although lexical distributional models of meaning, which represent the meaning of a word based on its context, taken from corpus co-occurrence statistics, are by now an essential part of the computational linguistics toolkit, its extension to larger phrases and sentences remains an issue that is starting to be explored.

In this paper we discuss compositionality from the perspective of VP ellipsis, which constitutes phrases that lack a verb phrase component that however is often \emph{marked} by another constituent in the phrase, often an auxiliary verb. Examples are \quotes{Mary drinks and Bob does too} and \quotes{Kim wears a hat but Sandy does not},  in either case the auxiliary verb \singlequotes{does} refers to the verb phrase occurring earlier in the sentence. VP ellipsis provides an interesting challenge  for compositional distributional semantics for two reasons: first,  it is a \emph{non-linear phenomenon}, as the verb phrase is needed twice to parse the sentence and there is no straightforward linear algebraic  way  to deal with the reuse of resources.  Secondly, one can easily extend the current experimental datasets of the setting \cite{grefenstette2011experimental,kartsaklis2013prior} and compute with it: VP ellipsis   is only one step further from a simple transitive sentence. The examples above are both constructed using two subjects, an object, and a verb. This makes modelling VP ellipsis a suitable candidate for experimental evaluation\footnote{This work is currently under review and could therefore not be included here as of yet.}.

Ellipsis has been modelled  both as a syntactic and a semantic problem \cite{kempson20154} and here we approach it from the perspective of  categorical compositional distributional semantics \cite{coecke2010mathematical,coecke2013lambek}, in which the derivations of a typelogical grammar are interpreted as linear maps on vector spaces. The tight connection between syntax and semantics that one assumes in these models means that we will treat ellipsis as a phenomenon that requires controlled copying of resources in syntax. Thus, similar to the approaches of \cite{jager1998multi,morrill2015computational,morrill2016logic}, we argue for the use of controlled forms of copying in a type logical system to deal with ellipsis. We then define two compositional architectures: a quantum entangled semantics following the direct categorical modelling of \cite{coecke2013lambek} where we use Frobenius algebras to interpret the copying operations, and a classical semantics in which an intermediate step allows for classical copying by means of variable reuse in terms of a non-linear lambda calculus. \\
\indent This paper is structured as follows: in Section \ref{sec:background},  we discuss the problem of ellipsis and argue for the legitimacy of non-linearity in the syntactic process. We define an extension of the Lambek calculus to deal with resource reuse in syntax in Section \ref{sec:syntax}. In  Sections \ref{sec:frobenius_semantics} and  \ref{sec:classical_semantics}, respectively,  we instantiate this calculus to a categorical and a classical model.  We then apply these two different semantics to derive different readings of elliptical phrases with structural ambiguities  and conclude in Section \ref{sec:conclusion}.

\section{Background}
\label{sec:background}

Loosely following \cite{dalrymple1991ellipsis,jager2006anaphora} we define ellipsis as a phenomenon in which two phrases are parallel in structure, though one of the phrases is incomplete and requires material from the other phrase to be \emph{copied} in order to make sense. In the case of verb phrase ellipsis there is usually a \emph{marker} present that specifies that material needs to be copied and moved into place.
%\begin{enumerate}
%	\item Ellipsis, ambiguity, and non-linearity
%	\item Quantum and Classical non-linearity
%	\item Two architectures for compositional distributional semantics
%\end{enumerate}
%\subsection{Ellipsis and Non-Linearity}
In the example of verb phrase ellipsis in Equation \ref{ex:simple_ellipsis}, where the elided verb phrase is marked by the auxiliary verb. Ideally, sentence \ref{ex:simple_ellipsis}(a) is in a bidirectional entailment relation with \ref{ex:simple_ellipsis}(b), i.e. (a) entails (b) and (b) entails (a).

\begin{equation}\label{ex:simple_ellipsis}
\begin{tabular}{cll}
$a$ & \quotes{Alice drinks and Bill does too}\\
$b$ & \quotes{Alice drinks and Bill drinks}\\
\end{tabular}
\end{equation}
A more complicated example of ellipsis and anaphora, that induces an ambiguity, is the one in Equation \ref{ex:ellipsis_anaphora}, where the ambiguous phrase (a) has two readings (b) and (c).
\begin{equation}\label{ex:ellipsis_anaphora}
\begin{tabular}{cll}
$a$ & \quotes{Gary loves his code and Bill does too} & (ambiguous)\\
$b$ & \quotes{Gary loves Gary's code and Bill loves Gary's code} & (strict)\\
$c$ & \quotes{Gary loves Gary's code and Bill loves Bill's code} & (sloppy)\\
\end{tabular}
\end{equation}
In a formal semantics account, the first example could be analysed with the auxiliary verb as an identity function on the main verb of the sentence and an intersective meaning for the coordinator. Somehow the parts need to be appropriately combined to produce the reading (b) for sentence (a):

\begin{center}
\begin{tabular}{lll}
	\begin{tabular}{c}
		does too : $\abs{x}{x}$ \\
		and : $\abs{x}{\abs{y}{(x \wedge y)}}$
	\end{tabular}
	& should give & $\mathbf{drinks}(\mathbf{alice}) \wedge \mathbf{drinks}(\mathbf{bill})$
\end{tabular}
\end{center}
The second example would assume the same meaning for the coordinator and auxiliary but now the possessive pronoun \quotes{his} gets a more complicated term: $\abs{x}{\abs{y}{\mathbf{owns}(x,y)}}$. The analysis then somehow should derive two readings:
\begin{center}
\begin{tabular}{ll}
	$\mathbf{loves}(\mathbf{gary},x) \wedge \mathbf{owns}(\mathbf{gary},x) \wedge \mathbf{loves}(\mathbf{bill},x)$ & (strict) \\[0.5em]
	$\mathbf{loves}(\mathbf{gary},x) \wedge \mathbf{owns}(\mathbf{gary},x) \wedge \mathbf{loves}(\mathbf{bill},y) \wedge \mathbf{owns}(\mathbf{bill},y)$ & (sloppy)
\end{tabular}
\end{center}
There are three issues with these analyses that a distributional semantic treatment should address: first, one has to determine how individual word meanings are combined to form the phrase meaning (syntax). Second, function words such as the coordinator \singlequotes{and} need to be given a lexical meaning since they should not be addressed distributionally (lexical semantics). Thirdly, the semantic representations are all non-linear: the main verb is used twice in the first example, the noun phrases are consumed twice or thrice in the second example. Somehow a model needs to account for how these non-linearities are obtained (derivational semantics).

To address the first issue we define an extension of the Lambek calculus that allows for the copying of resources by means of a controlled proof-theoretic contraction rule. The lexical semantics of the coordinator (\singlequotes{and}) and the possessive pronoun (\singlequotes{his}) are given by using Frobenius algebras. This approach was shown to be fruitful in previous work \cite{kartsaklis2016coordination,sadrzadeh2013frobenius}. We discuss the third issue below.

\paragraph{Quantum versus Classical non-linearity} To set up a compositional distributional model that allows a certain non-linearity there is a choice to be made: to stay within the existing categorical framework \cite{coecke2010mathematical,coecke2013lambek},  we want to work with compact closed categories and their  concrete instantiation to the category of vector spaces and linear maps. The Frobenius algebras that have been used in previous work to deal with relative pronouns and coordination, can also be used to allow the copying of resources, however this will happen in a non-cartesian way in which the material that was copied, is entangled. The main Frobenius map that is used for relative pronouns by \cite{sadrzadeh2013frobenius,moortgat2017lexical} expresses element wise multiplication, but its dual map copies a vector by placing its values on the diagonal of a square matrix. In terms of a type signature this indeed multiplies the vector space on which the map is performed, but does not allow for the actual vector to be used in a non-entangled way. In fact, there is no linear map that can copy arbitrary vectors in the cartesian sense \cite{abramsky2009no,jacobs2011bases}. To see this for a concrete example, consider the phrase \quotes{Alice loves herself} with tensors $\mathbf{alice} = \sum\limits_{i} a_i \vec{v}_i$, and $\mathbf{loves} = \sum\limits_{jkl} c_{jkl} (\vec{v}_j \tensor \vec{s}_k \tensor \vec{v}_l)$. The interpretation of a classical semantics (left) differs from the result of using Frobenius algebras (right):

\begin{center}
\begin{tabular}{@{}c@{\hskip 3em}c@{}}\toprule
	\textbf{Classical} & \textbf{Frobenius} \\[0.5em]
$\mathbf{alice}_i \mathbf{loves}_{ijk} \mathbf{alice}_k = \sum\limits_{ijk} a_i c_{ijk} a_k \vec{s}_{j}$ & $\mathbf{alice}_i \mathbf{loves}_{iji} = \sum\limits_{ij} a_i a_i c_{iji} \vec{s}_j$	\\
\bottomrule
\end{tabular}
\end{center}

%$$\mathbf{alice}_i \mathbf{loves}_{ijk} \mathbf{alice}_k = \sum\limits_{ijk} a_i c_{ijk} a_k \vec{s}_{j}$$
%whereas the result of using Frobenius algebras is something else:
%$$\mathbf{alice}_i \mathbf{loves}_{iji} = \sum\limits_{ij} a_i a_i c_{iji} \vec{s}_j$$
\noindent Within the categorical framework we could derive the sentence meaning on the right, but not the one on the left. However, we can define a different compositional distributional model that allows for the cartesian copying to take place. Since we do not want to fix this choice in advance, we define both models and give a comparison in their treatment of VP ellipsis to see whether a classical non-linearity is preferred over am entangled non-linearity.
%In order to still have Cartesian copying behaviour in a tensor based model, we decompose the DisCoCat model of \cite{coecke2013lambek} into a two-step architecture: we define an extension of the Lambek Calculus that allows for controlled contraction, and proofs of grammaticality, as well as lexical entries are interpreted in a non-linear simply typed $\lambda$-calculus. The second stage of interpretation interprets terms in a lambda calculus that abstractly models tensors, and their operations. The effect is that we allow the cartesian behavior of copying elements before concretisation in a vector semantics: the meaning of a sentence now is a program that has non-linear access to word embeddings.

\paragraph{Two Architectures} The categorical framework implements compositionality directly as a functorial passage from a syntactic category to a semantic category. The concrete model of \cite{coecke2013lambek} takes the Lambek calculus as a monoidal biclosed category, which then is mapped onto the category of vector spaces and linear maps, where the tensor product is monoidal and whose internal hom is given by the space of maps between two vector spaces. For our case, we take an extension of the Lambek calculus with controlled contraction which we will denote by $\mathbf{L_{\fdia,F}}$, and map it onto the category of vector spaces where each space has a Frobenius algebra, written $\mathbf{FVec_{Frob}}$. In a picture, the process looks like
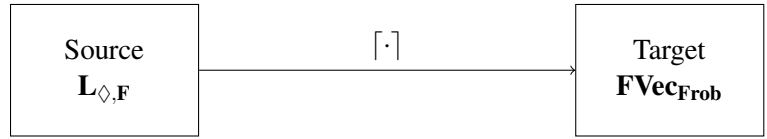
\begin{center}
\begin{tikzpicture}
	\node[draw, minimum width=2.5cm, minimum height=1.75cm] (der) 
	{\begin{tabular}{c}
	Source\\
	% \\
	$\mathbf{L_{\fdia,F}}$\\
	\end{tabular}};
	\node[draw, right=5cm of der, minimum width=2.5cm, minimum height=1.75cm] (sem)
	{\begin{tabular}{c}
	Target\\
	% \\
	$\mathbf{FVec_{Frob}}$\\
	\end{tabular}};
	\draw[->] (der) edge node[above] {$\F{\cdot}$} (sem);			
\end{tikzpicture}
\end{center}
%
%In a very general setting, compositionality can be defined as a homomorphic image (or functorial passage) from a syntactic algebra (or category) to a semantic algebra (or category). The only condition, then, is that the semantic algebra be weaker than the syntactic algebra: each syntactic operation needs to be interpretable by a semantic operation. To give a formal semantic account one would map the proof terms of a categorial grammar, or rewritings of a generative grammar, to the semantic operations of abstraction and application of some lambda term calculus. In a distributional model like the one of Coecke et al. \cite{coecke2010mathematical,coecke2013lambek}, derivations of a Lambek grammar are interpreted by linear maps on finite dimensional vector spaces. For our presentation it will suffice to say that the Lambek Calculus can be considered as a \emph{monoidal biclosed category}, which makes the mapping to the compact closure of vector spaces straightforward. However, we want to employ the copying power of non-linear lambda calculus, and so we will move from the direct interpretation below
In order to allow classical copying behaviour in a compositional distributional model, we decompose the categorical model of \cite{coecke2013lambek} into a two-step architecture: derivations and are now mapped onto terms of a non-linear simply typed lambda calculus $\mathbf{\lambda}_{NL}$. The second stage of the interpretation process replaces the assumed lexical constants for words by their lexical semantics, finally resulting in a term of a lambda calculus that models vectors and linear maps, denoted $\mathbf{\lambda_{FVec_{Frob}}}$. In a picture:
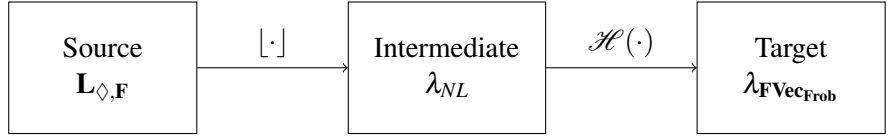
\begin{center}
\begin{tikzpicture}
	\node[draw, minimum width=2.5cm, minimum height=1.75cm] (der) 
	{\begin{tabular}{c}
	Source\\
	% \\
	$\mathbf{L_{\fdia, F}}$\\
	\end{tabular}};
	\node[draw, right=2cm of der, minimum width=2.5cm, minimum height=1.75cm] (sem1)
	{\begin{tabular}{c}
	Intermediate\\
	% \\
	$\mathbf{\lambda}_{NL}$\\
	\end{tabular}};
	\node[draw, right=2cm of sem1, minimum width=2.5cm, minimum height=1.75cm] (sem2)
	{\begin{tabular}{c}
	Target\\
	% \\
	$\mathbf{\lambda_{FVec_{Frob}}}$\\
	\end{tabular}};
	\draw[->] (der) edge node[above] {$\G{\cdot}$} (sem1);
	\draw[->] (sem1) edge node[above] {$\cal{H}(\cdot)$} (sem2);			
\end{tikzpicture}
\end{center}
The effect is that we allow the cartesian behaviour of copying elements before concretisation in a vector semantics: the meaning of a sentence now is a program that has non-linear access to word embeddings. \\
%What we end up with is in fact a more intricate target than in the direct case: target expressions are now lambda terms with a tensorial interpretation, i.e. a program with access to word embeddings. The next section gives all the details: we consider the syntax of the Lambek Calculus and our extension, the semantics of non-linear lambda calculus, and the interpretation of lambda terms in a lambda calculus with tensors.

\section{A Proof System for Controlled Copying}
\label{sec:syntax}
%The goal of this paper is to compare two different ways of composing distributional word embeddings. Although there are many proposals of typelogical grammars to deal with various linguistics phenomena, they very often compile away structural rules in the proof system they use, which counterfeits the categorical view of the framework of \cite{coecke2013lambek}. In this paper we therefore give a proof system that extends the Lambek calculus with unary modalities, solely used to control copying and movement, but leaving the categorical presentation of the proof system intact. In this way, the quantum entangled semantics that we want to derive can be defined in a straightforward way. \\

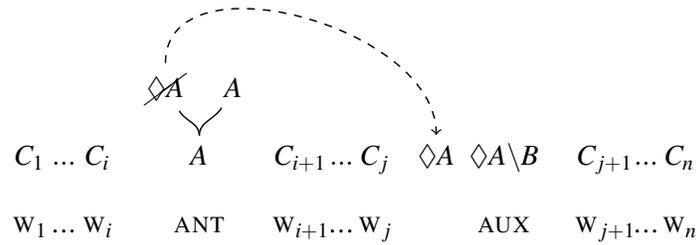
\begin{figure}[h]
\begin{center}
\begin{tikzpicture}[every node/.style={anchor=base},xscale=.9,yscale=.9]

\node (w1) at (1,0) {$\textsc{w}_1$};
\node (w2) at (1.5,0) {$\textsc{...}$};
\node (w3) at (2,0) {$\textsc{w}_i$};
\node (copy1) at (3,0) {};
\node (w4) at (3.5,0) {$\textsc{ant}$};
\node (copy2) at (4,0) {};
\node (w5) at (5,0) {$\textsc{w}_{i+1}$};
\node (w6) at (5.6,0) {$\textsc{...}$};
\node (w7) at (6.1,0) {$\textsc{w}_j$};
\node (gap) at (7,0) {};
\node (w8) at (8,0) {$\textsc{aux}$};
\node (w9) at (9.5,0) {$\textsc{w}_{j+1}$};
\node (w10) at (10.1, 0) {$\textsc{...}$};
\node (w11) at (10.6, 0) {$\textsc{w}_n$};

\node (t1) at (1,1) {$C_1$};
\node (t2) at (1.5,1) {$...$};
\node (t3) at (2,1) {$C_i$};

\node (t4) at (3.5,1) {$A$};
\node (tc1) at (3,2) {\cancel{$\Diamond A$}};
\node (tc2) at (4,2) {$A$};

\node (t5) at (5,1) {$C_{i+1}$};
\node (t6) at (5.6,1) {$...$};
\node (t7) at (6.1,1) {$C_j$};

\node (gapt) at (7,1) {$\Diamond A$};
\node (t8) at (8,1) {$\Diamond A \bs B$};

\node (t9) at (9.5, 1) {$C_{j+1}$};
\node (t10) at (10.1, 1) {$...$};
\node (t11) at (10.6, 1) {$C_n$};

\draw[->, line width=0.05em, dashed] (tc1) edge [out=90, in=90] (gapt);
\draw[line width=0.05em] (t4) edge [out=90,in=300] (tc1);
\draw[line width=0.05em] (t4) edge [out=90,in=240] (tc2);
%\node (n0) at (0,0) {\textsf{N}};
%\node (n1) at (1.3,0) {\textsf{N}};
%\node (t1) at (1.7,0.02) {$\tensor$};
%\node (n2) at (2.1,0) {\textsf{N}};
%\node (t2) at (2.5,0.02) {$\tensor$};
%\node (n3) at (2.9,0) {\textsf{N}};
%\node (t3) at (3.3,0.02) {$\tensor$};
%\node (n4) at (3.7,0) {\textsf{S}};
%\node (n5) at (5,0) {\textsf{N}};
%\node (n6) at (6.4,0) {\textsf{N}};
%\node (t4) at (6.8,0.02) {$\tensor$};
%\node (n7) at (7.2,0) {\textsf{N}};
%\node (t5) at (7.6,0.02) {$\tensor$};
%\node (n8) at (8,0) {\textsf{S}};
%
%\node (n9) at (2.1,2) {};
%\node (n10) at (2.1,-1.5) {};
%
%\draw (n0) edge [bend left=90] node[above=2pt] {$i$} (n1);
%\draw (n3) edge [bend left=90,pos=.2] node[above=2pt] {$k$} (n6);
%\draw (n4) edge [bend left=90, pos=.7] node[above=2pt] {$l$} (n8);
%\draw (n5) edge [bend left=90, pos=.8] node[above=2pt] {$m$} (n7);
%\draw (n0) edge [bend right=90]  node[below=2pt] {$i$} (n1);
%\draw (n3) edge [bend right=100, pos=.2] node[below=2pt] {$k$} (n7);
%\draw (n4) edge [bend right=90,pos=.7] node[below=2pt] {$l$} (n8);
%\draw (n5) edge [bend right=90] node[below=2pt] {$m$} (n6);
%
%\draw (n2) edge node[right] {$j$} (n9);
%\draw (n2) edge node[right] {$j$} (n10);

\end{tikzpicture}
\caption{General strategy for ellipsis resolution in $\mathbf{L_{\fdia, F}}$. The antecedent is copied, and the $\Diamond$ decorated copy is moved directly left of the marker which consumes the copy.}
\label{fig:matching}
\end{center}
\end{figure}

Our starting point for syntax is the Lambek Calculus, the noncommutative fragment of multiplicative linear logic without units. Formulas in $F$ are built from a set of basic formulas $B$ and using the connectives $\tensor, \bs, \s$, sharing the residuation relation\footnote{Algebraically, three operations $\cdot, \leftarrow, \rightarrow$ on a partial order form a residuated triple iff $a \leq c \leftarrow b \Leftrightarrow a \cdot b \leq c \Leftrightarrow b \leq a \rightarrow c$. Logically, this corresponds to the two-way inference rules that we use here.} expressed in Figure \ref{nldia}. Moreover, we add the control modalities $\Diamond, \Box$. The modalities have a purely syntactical role: instead of directly allowing the copying of resources, the system is designed such that only a type that is labelled with a $\Diamond$ can be copied. This prevents the overgeneration of a general contraction rule, but allows the copying of those words that have been decorated with a $\Diamond$ type. The residuated $\Box$ modality allows for the system to operate on the rest of a $\Diamond$ decorated type without losing track of the position of the $\Diamond$. Combining $\Diamond$ with a $\bs$ or $\s$ creates the behaviour wanted for ellipsis: an ellipsis marker will generally be annotated with a type $\Diamond A \bs B$, meaning that it expects a copy of a resource of type $A$ \emph{somewhere} to its left. Once the copy is created, it is moved into the right position to be consumed by the ellipsis marker, as expressed in Figure \ref{fig:matching}. To this end, the modalities license access to (limited forms of) contraction and commutativity, through the use of structural rules (see Figure \ref{nldia}). A similar setup has shown to give an account for pronoun relativisation \cite{moortgat1996multimodal}, which in the context of distributional semantics has been worked out in \cite{sadrzadeh2013frobenius,moortgat2017lexical}.

% EXAMPLE

% END EXAMPLE

\begin{figure}[h]
% identity, composition
\[\infer[]{\arr{1_A}{A}{A}}{}
\qquad
\infer[]{\arr{g\circ f}{A}{C}}{\arr{f}{A}{B} & \arr{g}{B}{C}}\]

% residuation
\[\infer[]{\arr{\resdia f}{A}{\gbox B}}{\arr{f}{\fdia A}{B}}
\qquad
\infer[]{\arr{\resright f}{A}{C/B}}{\arr{f}{A\tensor B}{C}}
\qquad
\infer[]{\arr{\resleft f}{B}{A\bs C}}{\arr{f}{A\tensor B}{C}}\]

\[\infer[]{\arr{\resdiainv g}{\fdia A}{B}}{\arr{g}{A}{\gbox B}}
\qquad
\infer[]{\arr{\resrightinv g}{A\tensor B}{C}}{\arr{g}{A}{C/B}}
\qquad
\infer[]{\arr{\resleftinv g}{A\tensor B}{C}}{\arr{g}{B}{A\bs C}}\]

%\[\arr{\xlefta}{A \tensor (B \tensor C)}{(A \tensor B) \tensor C}
%\qquad
%\arr{\xrighta}{(A \tensor B) \tensor C}{A \tensor( B \tensor C)}\]
\[\infer[]{\arr{\Xlefta(f)}{A \tensor(B \tensor C)}{D}}{\arr{f}{(A \tensor B) \tensor C}{D}}
\qquad
\infer[]{\arr{\Xrighta(f)}{(A \tensor B) \tensor C}{D}}{\arr{f}{A \tensor (B \tensor C)}{D}}\]
%\[\arr{\xlefta}{\fdia A\tensor(B\tensor C)}{(\fdia A\tensor B)\tensor C}
%\qquad
%\arr{\xrighta}{(A\tensor B)\tensor\fdia C}{A\tensor(B\tensor\fdia C)}\]
%
%\[\arr{\xleftc}{\fdia A\tensor(B\tensor C)}{B\tensor(\fdia A\tensor C)}
%\qquad
%\arr{\xrightc}{(A\tensor B)\tensor\fdia C}{(A\tensor\fdia C)\tensor B}\]
%\[\arr{C}{\fdia A}{\fdia A \tensor A}
%\qquad
%\arr{M}{(\fdia B \tensor A)\tensor C}{A \tensor (\fdia B \tensor C)}\]
%
%\[\arr{S}{\fdia B \tensor (\fdia A \tensor C)}{\fdia A \tensor (\fdia B \tensor C)}\]

\[\infer[]{\arr{\widehat{C}(f)}{A}{B}}{\arr{f}{\fdia A \tensor A}{B}}
\qquad
\infer[]{\arr{\widehat{M}(f)}{(\fdia B \tensor A)\tensor C}{D}}{\arr{f}{A \tensor (\fdia B \tensor C)}{D}}
\qquad
\infer[]{\arr{\widehat{S}(f)}{\fdia B \tensor (\fdia A \tensor C)}{D}}{\arr{f}{\fdia A \tensor (\fdia B \tensor C)}{D}}\]
\caption{\textbf{L}$_{\diamond,F}$. Residuation rules; Structural postulates for controlled copying and moving (rule form). The names of the rules are given by the term symbols in the conclusion of each rule.}
\label{nldia}
\end{figure}
The structural rules can also be stipulated in equivalent axiomatic form, but for the purpose of parsing it is more useful to consider rule form. We want to have a system that enjoys decidability: though this is not a straightforward property in the presence of just contraction\footnote{See the discussion in Katalin Bimb\'{o}'s monograph \cite{bimbo2014proof} and the results of \cite{surarso1996cut,chvalovsky2016full}.}, our system becomes decidable easily if we put a bound on the number of contractions in a proof.
%\begin{figure}[t]
%\[\arr{C}{\fdia A}{\fdia A \tensor A}
%\qquad
%\arr{M}{(\fdia B \tensor A)\tensor C}{A \tensor (\fdia B \tensor C)}\]
%%
%\[\arr{S}{\fdia B \tensor (\fdia A \tensor C)}{\fdia A \tensor (\fdia B \tensor C)}\]
%
%\caption{\textbf{NL}$_{\diamond,E}$. Structural postulates for controlled copying and moving.}
%\label{nlcopy}
%\end{figure}
%
%\begin{figure}[t]
%\[\infer[]{\arr{\widehat{C}(f)}{\fdia A}{B}}{\arr{f}{\fdia A \tensor A}{B}}
%\qquad
%\infer[]{\arr{\widehat{M}(f)}{(\fdia B \tensor A)\tensor C}{D}}{\arr{f}{A \tensor (\fdia B \tensor C)}{D}}
%\qquad
%\infer[]{\arr{\widehat{S}(f)}{\fdia B \tensor (\fdia A \tensor C)}{D}}{\arr{f}{\fdia A \tensor (\fdia B \tensor C)}{D}}\]
%
%\caption{\textbf{NL}$_{\diamond,E}$. Structural postulates for controlled copying and moving (rule form).}
%\label{nlcopyrule}
%\end{figure}

In order to determine the analysis of a phrase, we have a mapping $\sigma : \Sigma \rightarrow F$ where $\Sigma$ is a list of words. Sentencehood of a sequence of words $w_1,...,w_n$ is then determined by derivability with respect to a distinguished sentence type $s$ of the conjoined formula $\sigma(w_1) \tensor ... \tensor \sigma(w_n)$, and this generalises to arbitrary goal formulas. In other words, whenever $\sigma(w_1) \tensor ... \tensor \sigma(w_n) \rightarrow A$ is derivable, we may say that $w_1,...,w_n$ is a phrase \emph{of type} $A$.
\subsection{Deriving Ellipsis}

Let's consider again the examples given in Section \ref{sec:background}. An elliptical phrase like \quotes{Alice drinks and Bob does too} requires the verb phrase to be used twice; the proof system handles this by means of the structural rules of controlled contraction and movement. Figure \ref{fig:simpleder} shows the short-hand derivation for the phrase --- skipping trivial applications of associativity --- with the copy of the verb phrase highlighted in red. The derivation follows the general pattern depicted in Figure \ref{fig:matching}, with the auxiliary verb \quotes{does too} marking the ellipsis site and therefore typed $\fdia (np \bs s) \bs (np \bs s)$; requiring a copied verb from \emph{somewhere} to its left, it will return a verb. Reading the proof from bottom to top, first a contraction is applied to copy the verb phrase, marking the copy with a control modality. Then, the copy of the verb phrase is structurally moved rightward until it is in the right position to interact with the auxiliary ellipsis marker. It is not hard to see, then, that this allows the meaning of the verb phrase to be multiplied, have the copied version interact with the auxiliary to give the meaning of the whole phrase in a similar way as the meaning of the expanded phrase \quotes{Alice drinks and Bob drinks} would be computed.

\begin{figure}[h]
  \def\SUBPROOF{%
  \def\SUBPROOFONE{%
  \input{derivations/itv_app}%
}%
  \def\SUBPROOFTWO{%
  \input{derivations/itv_vp_site}%
}%
  \input{derivations/coordinate_phase}%
}%
  \input{derivations/struct_phase}%

\caption{Short hand derivation for \quotes{Alice drinks and Bob does-too}. The copied verb and its subformulas are highlighted in red. The bottom inferences copy the verb and move it to the auxiliary verb (structural phase), after which logical rules are applied to reduce to axiom leaves (logical phase).}
\label{fig:simpleder}
\end{figure}
\noindent More complicated cases of ellipsis contain anaphora, and allow an interaction that leads to strict and sloppy readings. To analyse the examples from Equation \ref{ex:simple_ellipsis}, the given proof system deals with the ambiguity by allowing a choice of resolution: either the anaphora gets resolved first, after which the resolved verb phrase is copied (strict reading), or the unresolved verb phrase is copied and resolved once for each subclause (sloppy reading). The two readings are derived in Figures \ref{fig:subproofs}, \ref{fig:sloppy}, \ref{fig:strict}. In the following two sections, we give respectively a quantum and a classical interpretation to these derivations.

\section{Quantum Semantics with Frobenius Algebras}
\label{sec:frobenius_semantics}

In order to derive the semantics for coordination and ellipsis proposed in \cite{kartsaklis2016coordination,kartsaklisverb}, we follow the framework of \cite{coecke2013lambek} and label the rules of our proof system with categorical morphisms in a compact closed category endowed with Frobenius algebras. Recall that a \emph{compact closed category} (CCC) is a monoidal category, i.e.~it has an associative $\tensor$ with unit $I$, and for every object there is a left and a right adjoint with maps
\[A^{l}\tensor A\Arrow{\epsilon^{l}} I \Arrow{\eta^{l}} A\tensor A^{l}
\qquad
A\tensor A^{r}\Arrow{\epsilon^{r}} I \Arrow{\eta^{r}} A^{r}\tensor A\]
These need satisfy the so-called yanking equations
\[ (1_A \tensor \epsilon^l_A) \comp (\eta^l_A \tensor 1_A) = 1_A \qquad (\epsilon^r_A \tensor 1_A) \comp (1_A \tensor \eta^r_A) = 1_A\]
\[ (\epsilon^l_A \tensor 1_{A^l}) \comp (1_{A^l} \tensor \eta^l_A) = 1_{A^l} \qquad (1_{A^r} \tensor \epsilon^r_A) \comp (\eta^r_A \tensor 1_{A^r}) = 1_{A^r}\]
In a \emph{symmetric} CCC, the tensor moreover is commutative, and we can write $A^{*}$ for the collapsed left and right adjoints. Frobenius algebras are defined over the objects of a category, and we say that an object $A$ has a Frobenius structure when it has maps
\begin{center}
\begin{tabular}{cccc}
	$\Delta : A \pijl A \tensor A$ & $\mu : A \tensor A \pijl A$ & $\iota: A \pijl I$ &  $\zeta : I \pijl A$ \\
\end{tabular}
\end{center}
satisfying monoidality and comonoidality for the pairs $(A, \mu, \zeta)$ and $(A, \Delta, \iota)$ and the Frobenius equation
$$ (\mu \tensor id_A) \comp (id_A \tensor \delta)  = \delta \comp \mu = (id_A \tensor \mu) \comp (\delta \tensor id_A)$$
In the concrete instance of \textbf{FVect}, the unit $I$ stands for the field $\mathbb{R}$; identity maps, composition and tensor product are defined as usual. Since bases of vector spaces are fixed in concrete models, there is only one natural way of defining a basis for a \emph{dual space}, so that $V^{*} \cong V$. In concrete models we may collapse the adjoints completely. The $\epsilon$ map takes inner products, whereas the $\eta$ map (with $\lambda = 1$) introduces an identity tensor as follows:
\begin{center}
	\begin{tabular}{l@{\hskip 1em}c@{\hskip 4em}ccl}
			$ \epsilon_V : V \tensor V \pijl \reals$ & given by & $\sum\limits_{ij} v_{ij} (\vec{e}_i \tensor \vec{e}_j)$ & $\mapsto$ & $\sum\limits_{i} v_{ii}$ \\[1.5em]
			$ \eta_V : \reals \pijl V \tensor V$ & given by & $\lambda$ & $\mapsto$ & $\sum\limits_i \lambda (\vec{e}_i \tensor \vec{e}_i) $
	\end{tabular}
\end{center}
Any finite vector space with fixed basis possesses a Frobenius structure, and so we write $\mathbf{FVect_{Frob}}$ for the category of finite dimensional vector spaces with fixed basis\footnote{In concrete experimental models bases are always fixed.}. The Frobenius maps take the form given below: $\Delta$ takes a tensor and places its values on the diagonal of a square matrix, whereas $\mu$ extracts the diagonal from a square matrix.
The $\iota$ and $\zeta$ maps respectively sum the coefficients of a vector or introduce a vector with the value $1$ for all of its coefficients.
\begin{center}
	\begin{tabular}{l@{\hskip 1em}c@{\hskip 4em}ccl}
			$ \Delta_V : V \pijl V \tensor V$ & given by & $\sum\limits_i v_i \vec{e}_i$ & $\mapsto$ & $\sum\limits_i v_i (\vec{e}_i \tensor \vec{e}_i)$ \\[1em]
			$ \iota_V : V \pijl \reals$ & given by & $\sum\limits_i v_i \vec{e}_i$ & $\mapsto$ & $\sum\limits_i v_i $\\[1em]
			$ \mu_V : V \tensor V \pijl V$ & given by & $\sum\limits_{ij} v_{ij} (\vec{e}_i \tensor \vec{e}_j)$ & $\mapsto$ & $\sum\limits_{i} v_{ii} \vec{e}_i$ \\[1em]
			$ \zeta_V : \reals \pijl V$ & given by & $\lambda$ & $\mapsto$ & $\sum\limits_i \lambda \vec{e}_i$
	\end{tabular}
\end{center}
\subsection{Interpretation: Proofs and Morphisms}
\label{subsec:interpretation}
The interpretation of derivations as linear maps has two components: on the type level, formulas are associated with vector spaces. On the proof level, the abstract terms of a proof become operations on vector spaces that respect the type interpretation.
On basic types, the interpretation $\F{\cdot}$ assigns arbitrary vector spaces, on complex types we have
\[\F{A\tensor B}=\F{A}\tensor\F{B}
\quad
\F{A/B}=\F{A}\tensor\F{B}^{*}
\quad
\F{A\bs B}=\F{A}^{*}\tensor\F{B}
\quad
\F{\fdia A} = \F{\gbox A} = \F{A}\]
On the level of proofs, we follow the standard interpretation given in \cite{coecke2013lambek,wijnholds2014categorical} to interpret the basic logic of residuation, and add the interpretation of the control rules from Figure \ref{nldia}. The simplest interpretations are identity and composition: $\F{1_A}=1_{\F{A}}$, $\F{g\circ f}=\F{g}\circ\F{f}$.
For the residuation inferences, we take the map $\arr{\F{f}}{\F{A}\otimes\F{B}}{\F{C}}$ interpreting the premise, and define
\[\begin{array}{c}
\F{\resright f} = \F{A}\Arrow{1_{\F{A}}\tensor\eta_{\F{B}}}\F{A}\tensor\F{B}\tensor\F{B}^{*}\Arrow{\F{f}\otimes 1_{\F{B}^{*}}}\F{C}\tensor\F{B}^{*}\\[1em]
\F{\resleft f} = \F{B}\Arrow{\eta_{\F{A}}\otimes 1_{\F{B}}}\F{A}^{*}\tensor\F{A}\tensor\F{B}\Arrow{1_{\F{A}^{*}}\otimes\F{f}}\F{A}^{*}\tensor\F{C}
\end{array}\]
For the inverses, from maps $\arr{\F{g}}{\F{A}}{\F{C/B}}$, $\arr{\F{h}}{\F{B}}{\F{A\bs C}}$ for the premises, we define
\[\begin{array}{c}
\F{\resrightinv g} = \F{A}\tensor\F{B}\Arrow{\F{g}\tensor 1_{\F{B}}}\F{C}\tensor\F{B}^{*}\tensor\F{B}\Arrow{1_{\F{C}}\tensor\epsilon_{\F{B}}}\F{C}\\[1em]
\F{\resleftinv h} = \F{A}\tensor\F{B}\Arrow{1_{\F{A}}\tensor\F{h}}\F{A}\tensor\F{A}^{*}\tensor\F{C}\Arrow{\epsilon_{\F{A}}\tensor 1_{\F{C}}}\F{C}
\end{array}\]
For the (derived) rules of monotonicity, the case of parallel composition is immediate: $\F{f\tensor g}=\F{f}\tensor\F{g}$. For the slash cases,
from $\arr{\F{f}}{\F{A}}{\F{B}}$ and $\arr{\F{g}}{\F{C}}{\F{D}}$, we obtain

\begin{tikzpicture}
\matrix (m) [matrix of math nodes, row sep=3em, column sep=4em]
{\F{f/g}= & \F{f\bs g}= \\[-2em]
\F{A}\tensor\F{D}^{*} &  \F{B}^{*}\tensor\F{C} \\
\F{B}\tensor\F{C}^{*}\tensor\F{C}\tensor\F{D}^{*} & \F{B}^{*}\tensor\F{A}\tensor\F{A}^{*}\tensor\F{D} \\
\F{B}\tensor\F{C}^{*}\tensor\F{D}\tensor\F{D}^{*} & \F{B}^{*}\tensor\F{B}\tensor\F{A}^{*}\tensor\F{D} \\
\F{B}\tensor\F{C}^{*} & \F{A}^{*}\tensor\F{D} \\};

\path[-stealth] (m-2-1) edge node [left] {$\F{f}\tensor\eta_{\F{C}}\tensor 1_{\F{D}^{*}}$} (m-3-1);
\path[-stealth] (m-3-1) edge node [left] {$1_{\F{B}\tensor\F{C}^{*}}\tensor\F{g}\tensor 1_{\F{D}^{*}}$} (m-4-1);
\path[-stealth] (m-4-1) edge node [left] {$1_{\F{B}\tensor\F{C}^{*}}\tensor\epsilon_{\F{D}}$} (m-5-1);

\path[-stealth] (m-2-2) edge node [right] {$1_{\F{B}^{*}}\tensor\eta_{\F{A}}\tensor \F{g}$} (m-3-2);
\path[-stealth] (m-3-2) edge node [right] {$1_{\F{B}^{*}}\tensor\F{f}\tensor 1_{\F{A}^{*}\tensor\F{D}}$} (m-4-2);
\path[-stealth] (m-4-2) edge node [right] {$\epsilon_{\F{B}}\tensor 1_{\F{A}^{*}\tensor\F{D}}$} (m-5-2);
\end{tikzpicture}

\noindent
Interpretation for the associativity structural rules is immediate via the standard associativity of \textbf{FVect}: $\F{\Xlefta f} = \F{f} \comp \alpha^{-1}$ and $\F{\Xrighta f} = \alpha \comp \F{f}$. For the other structural rules, we additionally use the symmetry maps of \textbf{FVect} as well as the diagonal Frobenius map $\Delta$:
%\F{\resrightinv g} = \F{A}\tensor\F{B}\Arrow{\F{g}\tensor 1_{\F{B}}}\F{C}\tensor\F{B}^{*}\tensor\F{B}\Arrow{1_{\F{C}}\tensor\epsilon_{\F{B}}}\F{C}\\
%\F{\resleftinv h} = \F{A}\tensor\F{B}\Arrow{1_{\F{A}}\tensor\F{h}}\F{A}\tensor\F{A}^{*}\tensor\F{C}\Arrow{\epsilon_{\F{A}}\tensor 1_{\F{C}}}\F{C}
%\[\begin{array}{c}
\begin{center}
$\F{\widehat{C}(f)} = \F{A} \Arrow{\Delta_{\F{A}}} \F{A} \tensor \F{A} \Arrow{\F{f}} \F{B}$ \\[1em]
$\F{\widehat{M}(f)} = (\F{B} \tensor \F{A}) \tensor \F{C} \Arrow{\sigma_{\F{B},\F{A}} \tensor 1_{\F{C}}} (\F{A} \tensor \F{B}) \tensor \F{C} \Arrow{\alpha} \F{B} \tensor (\F{A} \tensor \F{C}) \Arrow{\F{f}} \F{D}$ \\[1em]
$\F{\widehat{S}(f)} = \F{B} \tensor (\F{A} \tensor \F{C}) \Arrow{\alpha^{-1}} (\F{B} \tensor \F{A}) \tensor \F{C} \Arrow{\sigma_{\F{B}, \F{A}} \tensor 1_{\F{C}}} (\F{A} \tensor \F{B}) \tensor \F{C} \Arrow{\F{f} \comp \alpha} D$\\
%$\F{\widehat{S}(f)} = \F{B} \tensor (\F{A} \tensor \F{C}) \Arrow{\alpha^{-1}} (\F{B} \tensor \F{A}) \tensor \F{C} \Arrow{\sigma_{\F{B}, \F{A}} \tensor 1_{\F{C}}} (\F{A} \tensor \F{B}) \tensor \F{C} \Arrow{\alpha} \F{A} \tensor (\F{B} \tensor \F{C}) \Arrow{\F{f}} D$\\
%\end{array}\]
%\begin{tikzpicture}
%\matrix (m) [matrix of math nodes, row sep=3em, column sep=4em]
%{\F{\widehat{M}(f)}= & \F{\widehat{S}(f)}= \\[-2em]
%(\F{B} \tensor \F{A}) \tensor \F{C} & \F{B} \tensor (\F{A} \tensor \F{C})\\
%(\F{A} \tensor \F{B}) \tensor \F{C} & (\F{B} \tensor \F{A}) \tensor \F{C}\\
%\F{A} \tensor (\F{B} \tensor \F{C}) & (\F{A} \tensor \F{B}) \tensor \F{C} \\
%D & \F{A} \tensor (\F{B} \tensor \F{C}) \\
%& D \\};
%%
%\path[-stealth] (m-2-1) edge node [left] {$\sigma_{\F{B},\F{A}} \tensor 1_{\F{C}}$} (m-3-1);
%\path[-stealth] (m-3-1) edge node [left] {$\alpha$} (m-4-1);
%\path[-stealth] (m-4-1) edge node [left] {$\F{f}$} (m-5-1);
%%
%\path[-stealth] (m-2-2) edge node [right] {$\alpha^{-1}$} (m-3-2);
%\path[-stealth] (m-3-2) edge node [right] {$\sigma_{\F{B}, \F{A}} \tensor 1_{\F{C}}$} (m-4-2);
%\path[-stealth] (m-4-2) edge node [right] {$\alpha$} (m-5-2);
%\path[-stealth] (m-5-2) edge node [right] {$\F{f}$} (m-6-2);
%\end{tikzpicture}
\end{center}

\subsection{Frobenius Semantics of Ellipsis}

Setting $\F{np} = N, \F{s} = S$, the proof in Figure \ref{fig:simpleder} will be mapped on a morphism 
$$N \tensor (N \tensor S) \tensor (S \tensor S \tensor S) \tensor N \tensor (N \tensor S \tensor N \tensor S) \Arrow{F} S$$
Let us write $[f]$ for a morphism which contains $f$ but is surrounded by identity morphisms, and let us index it $[f_{A_i}]$ to specify the map acts on the $i$th occurrence of the object $A$ in an object, needed whenever there may be an ambiguity. Then, we can write the full morphism as
$$F = [\epsilon_N \tensor \epsilon_S \tensor id_S \tensor \epsilon_S] \comp [\epsilon_{N_3}] \comp [\epsilon_{N_4 \tensor S_5}] \comp [\sigma_{N_3 \tensor S_5, N_4}] \comp [\sigma_{N_3 \tensor S_2, S_3 \tensor S_4 \tensor S_5}] \comp [\sigma_{N_2 \tensor S_1, N_3 \tensor S_2}] \comp [\Delta_{N_2 \tensor S_1}]$$
That is, the Frobenius copying map is applied to the content of the verb tensor, after which the \singlequotes{copy} is then moved to the right into the position next to the space $N \tensor S \tensor N \tensor S$ in which the auxiliary expression lives. We then perform contractions in the expected way. For concrete maps we will write $\times$ for the tensor contraction defined in the previous section, and $\cdot^{\bot}$ for the transpose of a tensor; $\odot$ denotes element wise multiplication. We can use boldface for concrete tensors to get the final concrete map
$$\mathbf{alice} \tensor \mathbf{drinks} \tensor \mathbf{and} \tensor \mathbf{bob} \tensor \mathbf{does \ too} \mapsto (\mathbf{alice} \odot \mathbf{bob})^{\bot} \times \mathbf{drinks}$$
And similarly, for a transitive case, one would obtain the meaning
	$$(\mathbf{alice} \odot \mathbf{bob})^{\bot} \times (\mathbf{drinks} \times \mathbf{beer})$$
For the more involved examples of Figures \ref{fig:sloppy}, \ref{fig:strict} we obtain more complicated maps that make a choice in the order of resolution of the ellipsis and the anaphoric reference. Rewriting the obtained linear map using the \singlequotes{spider} equation of a commutative special Frobenius algebra, the normal form for the strict derivation, in which Gary fulfills the role of the anaphora before the ellipsis is resolved, gives
	$$\mathbf{gary} \tensor \mathbf{loves} \tensor \mathbf{his} \tensor \mathbf{code} \tensor \mathbf{and} \tensor \mathbf{bob} \tensor \mathbf{does \ too} \mapsto \mu_N(\mathbf{gary} \odot \mathbf{bob} \odot \mathbf{code})_{ik} \mathbf{loves}_{ijk}$$
That is, in the final result the two subjects and the object are multiplied element wise, after which the resulting vector is expanded to be consumed by the verb. Note that the meaning of the phrase \quotes{Gary loves Gary's code and Bob loves Gary's code} is thus the same as the meaning of \quotes{Gary loves Bob's code and Bob loves Bob's code}, due to the fact that we instantiated the anaphoric element \singlequotes{his} with element wise multiplication.

For the sloppy derivation (\quotes{Gary loves Gary's code and Bob loves Bob's code}) the situation is more problematic; the normal form for the meaning of the sloppy phrase is identical to the strict reading:
	$$\mathbf{gary} \tensor \mathbf{loves} \tensor \mathbf{his} \tensor \mathbf{code} \tensor \mathbf{and} \tensor \mathbf{bob} \tensor \mathbf{does \ too} \mapsto \mu_N(\mathbf{gary} \odot \mathbf{bob} \odot \mathbf{code})_{ik} \mathbf{loves}_{ijk}$$
So not only do the readings have a symmetric meaning, the two different readings get the same semantics. The only way to mend this would be to change the lexical meaning of the anaphora \singlequotes{his} for an operator other than element wise multiplication, though we suspect similar problems will arise.
%$$ id_N \tensor \Delta_{N \tensor S} \tensor id_{S \tensor S \tensor S \tensor N \tensor N \tensor S \tensor N \tensor S}$$
%$$N \tensor (N \tensor S) \tensor (N \tensor S) \tensor (S \tensor S \tensor S) \tensor N \tensor (N \tensor S \tensor N \tensor S)$$
%$$[\sigma_{N \tensor S, N}] \comp [\sigma_{N \tensor S, S \tensor S \tensor S}] \comp [\sigma_{N \tensor S, N \tensor S}]$$
%$$N \tensor (N \tensor S) \tensor (S \tensor S \tensor S) \tensor N \tensor (N \tensor S) \tensor (N \tensor S \tensor N \tensor S)$$
%$$[\epsilon_N \tensor \epsilon_S \tensor id_S \tensor \epsilon_S] \comp [\epsilon_N] \comp [\epsilon_{N \tensor S}]$$
%$$S$$

\section{Classical Semantics with Lambdas and Tensors}
\label{sec:classical_semantics}

In this section,  we develop a two-step semantics:  first, we map the sentences to   a non-linear lambda calculus,  and from there to vectors and tensors. The meaning of a sentence now is a program with non-linear access to word embeddings. As depicted in the diagrammatic plan in Section \ref{sec:background}, we deploy the same proof theory as in the one-step setup above, but rather than mapping types to vector spaces and proofs to linear maps, we map respectively to types and terms of a non-linear lambda calculus, after which a second translation steps replaces lexical constants associated with words by (terms modelling) concrete vectors and linear maps. In the next two subsections we work out these two steps.

\subsection{Derivational Semantics: Lambdas and Constants}
\label{subsec:deriv_lambda}

In the first step of the interpretation process, we map types and proofs onto types and terms of a non-linear simply typed lambda calculus with products. Similar to subsection \ref{subsec:interpretation}, we define an arbitrary interpretation on basic types $\G{.}$, and define on complex types
\[\G{A\tensor B}=\G{A} \times \G{B}
\quad
\G{A/B}=\G{A} \rightarrow \G{B}
\quad
\G{A\bs B}=\G{A} \rightarrow \G{B}
\quad
\G{\fdia A} = \G{\gbox A} = \G{A}\]
For the proofs, we can interpret identity and composition straightforwardly by $\lambda x. x$ and $\lambda x. N \ (M \  x)$ for $M$ and $N$ the terms of the subproofs, respectively. The binary residuation rules correspond to application and abstraction depending on the direction in which the rule is applied, whereas unary residuation does not change the terms at all
\[\begin{array}{l@{\hskip 3em}l@{\hskip 3em}l}
\G{\resright M} = \lambda x \ y. M \ \langle x, y \rangle & \G{\resleft M} = \lambda y \ x. M \ \langle x, y \rangle  & \G{\resdia M} = M \\[0.5em]
\G{\resrightinv N} = \lambda \langle x, y \rangle. (N \ x) \ y & \G{\resleftinv N} = \lambda \langle x, y \rangle. (N \ y) \ x  & \G{\resdiainv N} = N \\
\end{array}\]
The derived monotonicity rules get the interpretation below:
\[\begin{array}{l@{\hskip 3em}l@{\hskip 3em}l}
\G{M \tensor N} = \lambda \langle x, y \rangle. \langle M \ x, N \ y \rangle & \G{M \bs N} = \lambda f \ x. N \ (f \ (M \ x))& \G{M \s N} = \lambda f \ x. M \ (f \ (N \ x)) \\
\end{array}\]
The associativity rules behave as an identity since associativity is implicit in lambda terms.
\[\begin{array}{l@{\hskip 3em}l}
\G{\Xlefta(M)} = \lambda \langle x, y, z \rangle. M \ \langle x, y, z \rangle & \G{\Xrighta(M)} = \lambda \langle x, y, z \rangle. M \ \langle x, y, z \rangle \\
\end{array}\]
The non-linear behaviour enters with the interpretation for the structural rules for movement and copying:
\[\begin{array}{l@{\hskip 3em}l@{\hskip 3em}l}
\G{\widehat{C}(M)} = \lambda x. M \ \langle x, x \rangle & \G{\widehat{M}(M)} = \lambda \langle y, x, z \rangle. M \ \langle x, y, z \rangle & \G{\widehat{S}(M)} =  \lambda \langle y, x, z \rangle. M \ \langle x, y, z \rangle \\
\end{array}\]
In this first step of interpretation, the words of a phrase are assigned constants in a term that is built up by translating the proof term using above translation. For the sample derivation of Figure \ref{fig:simpleder}, the logical phase computes only reductions; the subproof of the type
$$(np_{} \tensor np_{} \bs s_{}) \tensor ((s_{} \bs s_{}) \s s_{} \tensor (np_{} \otimes  (\redbox{\Diamond (np_{} \bs s_{})} \otimes \ \Diamond (np_{} \bs s_{}) \bs  (np_{} \bs s_{})))) \longrightarrow s_{}$$
gives an abstract term
$$\lambda \langle \texttt{subj}_1, \texttt{verb}, \texttt{coord}, \texttt{subj}_2, \texttt{verb}^{\ast}, \texttt{aux} \rangle. (\texttt{coord} \ ((\texttt{aux} \ \texttt{verb}^{\ast}) \ \texttt{subj}_2)) (\texttt{verb} \ \texttt{subj}_1)$$
and the structural phase repositions the copy $\texttt{verb}^{\ast}$ next to the verb, after which the contraction rule \emph{identifies} the variables associated with them, unifying $\texttt{verb}$ and $\texttt{verb}^{\ast}$:
$$\lambda \langle \texttt{subj}_1, \texttt{verb}, \texttt{coord}, \texttt{subj}_2, \texttt{aux} \rangle. (\texttt{coord} \ ((\texttt{aux} \ \texttt{verb}) \ \texttt{subj}_2)) (\texttt{verb} \ \texttt{subj}_1)$$
We get the final \emph{abstract proof term} for the proof in Figure \ref{fig:simpleder} by applying the term above to the constants for the words in the sentence:
\begin{equation}
(\texttt{and} \ ((\texttt{dt} \ \texttt{drinks}) \ \texttt{bob})) (\texttt{drinks} \ \texttt{alice}) : s
\label{eq:abstract}
\end{equation}

\subsection{Lexical Semantics: Lambdas and Tensors}
\label{subsec:interpretation}

We complete the vector semantics by adding the second step in the interpretation process, which is the insertion of lexical entries for the assumptions/constants occurring in a proof term. In this step we face the issue that interpretation directly into a vector space is not an option given that there is no copying map that is linear, while at the same time lambda terms don't seemingly reflect vectors. However, following \cite{muskens2016context} we can model vectors using a lambda calculus as shown in the subsection below.

\subsubsection{Lambdas and Tensors}

Vectors can be seen as functions from natural numbers to the values in the underlying field, allowing us to represent them naturally as lambda terms. For any dimensionality $n$, we assume a basic type $I_n$, representing a finite index set (in concrete models the number of index types will be finite). The underlying field, in our case the real numbers $\reals$, is given by the type $R$.

The type of a vector in $\reals^n$ is now $V^n = I_n \rightarrow R$, the type of an $n \times m$ matrix is $M^{n\times m} = I_n \rightarrow I_m \rightarrow R$. In general, we may represent an arbitrary tensor with dimensions $n,m,...,p$ by $T^{n\times m ... \times p} = I_n \rightarrow I_m \rightarrow ... \rightarrow I_p \rightarrow R$. We will leave out the superscripts denoting dimensionality when they are either irrelevant or understood from the context.

By reference to index notation for linear algebra, we write $v \ i$ as $v_i$ whenever it is understood that $i$ is of type $I$.
%
%\begin{center}
%\begin{tabular}{m{2em}m{5em}}
%$
%\begin{pmatrix}
%	4 \\
%	2 \\
%	7
%\end{pmatrix}
%$
%&
%$
%\begin{matrix}
%f : I_3 \rightarrow R \\
%f(1) = 4 \\
%f(2) = 2 \\
%f(3) = 7 \\
%\end{matrix}
%$
%\end{tabular}
%\end{center}
We moreover assume constants for the basic operations of a vector space: $0 : R, 1 : R, + : R \rightarrow R \rightarrow R, \cdot : R \rightarrow R \rightarrow R$ with their standard interpretation. Standard operations can now be expressed: \\

\begin{center}
\ra{2}
\begin{tabular}{@{}ccc@{}}\toprule
	Name & Symbol & Lambda term \\ \midrule
	Matrix transposition & $\cdot^T$ & $\lambda mij. m_{ji} : M \rightarrow M$ \\
	Matrix multiplication & $\times_1$ & $\lambda mvi. \sum\limits_{j} m_{ij} \cdot v_j : M \rightarrow V \rightarrow V$ \\
	Cube multiplication & $\times_2$ & $\lambda cvij. \sum\limits_{k} c_{ijk} \cdot v_k : C \rightarrow V \rightarrow M$ \\
	Element wise multiplication & $\odot$ & $\lambda uvi. u_i \cdot v_i : V \rightarrow V \rightarrow V$ \\
	\bottomrule
\end{tabular}
\end{center}
\ \\
%We can also express backwards matrix multiplication by composing matrix transposition with standard multiplication: $\times^T := \lambda mvi. \sum\limits_j m_{ji} \cdot v_j : M \rightarrow V \rightarrow V$. \\

\subsubsection{Lexical substitution}

To obtain a concrete model for a phrase, we need replace the constants $c$ in a proof term by their vectorial representation. This is done by means of a lexicon of semantic terms, that induces a homomorphism $\cal{H}$ on terms. Table \ref{table:homomorphism} below gives the substitutions for a contraction-based and an additive-multiplicative model respectively. Translating the abstract proof term from Equation \ref{eq:abstract} using the contraction-based model means reducing the term below to give the final, classical meaning:
\begin{table}[h]
\begin{center}
\ra{1.4}
\begin{tabular}{@{}lllll@{}}\toprule
$w$&$\sigma(w)$ & ${\cal H}_1(w)$ & ${\cal H}_2(w)$ & ${\cal T}(w)$\\ \midrule
\texttt{cn}& $n$ & \text{\bf cn}& \text{\bf cn} & $V$\\
\texttt{adj}& $np \s n$ & $\lambda v.(\text{\bf adj}\times_1 v)$ & $\lambda v.(\text{\bf adj} + v)$ & $VV$ \\
\texttt{adv}& $(np \bs s) \bs (np \bs s)$ & $\lambda M.M$ & $\lambda M.(\mathbf{adv} + M)$ & $(VV)VV$ \\
\texttt{itv}& $np \bs s$ & $\lambda v.(\text{\bf itv}\times_1 v)$ & $\lambda v.(\text{\bf itv} + v)$ & $VV$\\
\texttt{tv}& $(np \bs s) \s np$ & $\lambda uv.(\text{\bf tv}\times_2 v)\times_1 u$ & $\lambda uv.(\text{\bf tv} + v + u)$ & $VVV$\\
\texttt{coord}& $(s \bs s) \s s$ & $\lambda P.\lambda Q. P \odot  Q$ & $\lambda P.\lambda Q. (P \odot Q)$ & $VVV$\\
\texttt{quant} & $(s \s (np \bs s)) \s n$ & $\lambda vZ. Z(\mbox{\textbf{quant}}\times_1 v)$&$\lambda vZ. Z(\mbox{\textbf{quant}} + v)$&$V(VV)V$\\
\bottomrule
\end{tabular}
\end{center}
\caption{Translation that sends abstract terms to a tensor-based model using tensor contraction (column 3) or addition (column 4) as the main operation. In both models the coordinator is interpreted using element wise multiplication. Note that $\mathbf{adj}$, $\mathbf{itv}$, $\mathbf{tv}$ and $\mathbf{quant}$ denote vectors under ${\cal H}_2$.}
\label{table:homomorphism}
\end{table}

$$(\lambda P. \lambda Q. P \odot Q \ ((\lambda x.x \ (\lambda v. \mathbf{drinks} \times_1 v)) \ \textbf{bob})) ((\lambda v. (\mathbf{drinks} \times_1 v)) \ \textbf{alice})$$
$$\rightarrow_{\beta} (\lambda P. \lambda Q. P \odot Q \ ((\lambda v. \mathbf{drinks} \times_1 v) \ \textbf{bob})) ((\lambda v. (\mathbf{drinks} \times_1 v)) \ \textbf{alice})$$
$$\rightarrow_{\beta} (\lambda P. \lambda Q. P \odot Q \ (\mathbf{drinks} \times_1 \textbf{bob})) (\mathbf{drinks} \times_1 \textbf{alice})$$
$$\rightarrow_{\beta} (\mathbf{drinks} \times_1 \textbf{bob}) \odot (\mathbf{drinks} \times_1 \textbf{alice})$$
%Using the table above, we can translate the proof term of Figure \ref{fig:simpleder}
%$$(\texttt{and} \ ((\texttt{dt} \ \texttt{drinks}) \ \texttt{bob})) (\texttt{drinks} \ \texttt{alice}) : s$$
%and substitute the concrete terms to get
%$$\rightarrow_{\beta} (\mathbf{drinks} \times_1 \mathbf{alice}) \odot (\mathbf{drinks} \times_1 \mathbf{bob})$$
As an alternative, we can instantiate the multiplicative-additive model as well, in which the transitive sentences are obtained by adding the individual word embeddings, but the overall result is got by multiplying the two sentence vectors. The final meaning now is
$$(\mathbf{drinks} + \mathbf{alice}) \odot (\mathbf{drinks} + \mathbf{bob})$$
%\begin{table}[h]
%\begin{center}
%\ra{1.4}
%\begin{tabular}{llll}\toprule
%$w$&$\sigma(w)$ & ${\cal H}(w)$&${\cal T}(w)$\\ \midrule
%\texttt{cn}& $n$ & \text{\bf cn}&$V$\\
%\texttt{adj}& $np \s n$ & $\lambda v.(\text{\bf adj} + v)$&$VV$ \\
%\texttt{adv}& $(np \bs s) \bs (np \bs s)$ & $\lambda m.(\text{\bf adv} + m)$&$MM$ \\
%\texttt{itv}& $np \bs s$ & $\lambda v.(\text{\bf itv} + v)$& $VV$\\
%\texttt{tv}& $(np \bs s) \s np$ & $\lambda uv.(\text{\bf tv} + v + u)$&$VVV$\\
%\texttt{coord}& $(s \bs s) \s s$ & $\lambda P.\lambda Q. (P \odot Q)$&$VVV$\\
%\texttt{quant} & $(s \s (np \bs s)) \s n$ & $\lambda vZ. Z(\mbox{\textbf{quant}} + v)$&$V(VV)V$\\
%\bottomrule
%\end{tabular}
%\end{center}
%\caption{Translation that sends abstract terms to a multiplicative-additive model.}
%\end{table}
The general description for a simple elliptical phrase that comes out of this is
	$$M(\mathbf{alice}, \mathbf{drinks}) \ \nabla \ M(\mathbf{bob}, N(\mathbf{drinks}))$$
where $M$ is a general model for an intransitive sentence, and $N$ is a model that could potentially modify the verb tensor. Similarly, the recipe for a transitive elliptical phrase like \quotes{Alice drinks beer and Bob does too} will be
	$$M(\mathbf{alice}, \mathbf{drinks}, \mathbf{beer}) \ \nabla \ M(\mathbf{bob}, N(\mathbf{drinks}), \mathbf{beer})$$
For the sloppy/strict readings involving anaphora, we give the sloppy and strict interpretation in a contraction-based model of the sentence \quotes{Gary loves his code and Bill does too}:
\begin{center}
\begin{tabular}{lc}
$(\mathbf{gary} \times_1 \mathbf{loves} \times_2 (\mathbf{gary} \odot \mathbf{code})) \odot (\mathbf{bob} \times_1 \mathbf{loves} \times_2 (\mathbf{bob} \odot \mathbf{code}))$ & (strict) \\[1em]
	$(\mathbf{gary} \times_1 \mathbf{loves} \times_2 (\mathbf{gary} \odot \mathbf{code})) \odot (\mathbf{gary} \times_1 \mathbf{loves} \times_2 (\mathbf{bob} \odot \mathbf{code}))$ & (sloppy)
\end{tabular}
\end{center}
So we see that in a classical semantics we can distinguish the two readings and moreover they do not coincide with phrases in which subjects and objects are swapped.

\section{Discussion, Conclusion, Further Work}
\label{sec:conclusion}

%\textbf{Summary}\\
In this paper we incorporated a proper notion of copying into a compositional distributional model of meaning to deal with some selected cases of ellipsis and anaphora. We developed two different concrete vector semantics for an extension of the Lambek Calculus with control modalities that allow for the copying of resources: in the first semantics we followed the categorical framework of \cite{coecke2013lambek} and derived sentence meanings similar to the proposal of Kartsaklis \cite{kartsaklisverb}, but found that for ambiguous cases of ellipsis the meaning of the unambiguous interpretations coincide. In the second semantics, we took a two-step approach by translating proofs to terms of a non-linear lambda terms, that then get concretised in several different models. Here we retain the different semantics for different interpretations of ambiguous elliptical phrases. Some initial experiments have shown, however, that the two frameworks give comparable results in a similarity task involving ellipsis, suggesting that the entangled semantics serve as a good linear approximation of a non-linear phenomenon.\footnote{This work is currently under review and could therefore not be included as of yet.}.
% By decomposing the DisCoCat architecture into a two step interpretation process, we were able to combine the flexibility of the Cartesian structure of the non-linear simply typed lambda calculus, with a vector based representation of word meanings. We gave an analysis of elliptic phrases and showed how the elliptic phrases get assigned the same meaning as their resolved variants. We also carried out a simple experiment, based on previous results on sentence similarity to show that existing models can be easily extended to give good results on a small elliptic similarity task. \\
%\ \\
%\textbf{Discussion} \\
%
%\ \\
%\textbf{Future Work}\\
To complete this work in the future, we would like to carry out a large scale experiment to compare the linear approximative model of the categorical framework, and the classical semantics using non-linear lambda terms. Furthermore, we would like to experiment with the kind of derivational ambiguities that arise in the elliptical setting with anaphora.
%
%We have seen how a typelogical distributional semantics can be used to assign several meanings to the same surface form by means of derivational ambiguity. One open issue is how to validate in an experiment the effectiveness of a particular compositional model. We aim to explore this in future, more experimentally oriented work.

\paragraph{Acknowledgements}
The authors are thankful for the anonymous reviewers' comments. The first author gratefully acknowledges support from a Queen Mary Principal Studentship.

%\nocite{*}
\bibliographystyle{eptcs}
\newcommand{\references}{capns_wijnsadr}
\bibliography{\references}   % name your BibTeX data base

%\begin{thebibliography}{}
%%
%% and use \bibitem to create references. Consult the Instructions
%% for authors for reference list style.
%\end{thebibliography}

\newpage
\appendix

\begin{landscape}
\section{Proofs}
%\topskip0pt
%\vspace*{\fill}
\begin{center}
\begin{figure}[h]
\resizebox{\linewidth}{!}{
\begin{tabular}{cc}
  \input{derivations/gary_loves_gary_his_code}%
 & %
  \input{derivations/gary_loves_his_code}%
 \\[1em]
$\lpair{x}{\lpair{y}{\lpair{\lpair{z}{u}}{v}}} : \leftapp{\rightapp{y}{\rightapp{\leftapp{u}{z}}{v}}}{x}$ & $\lpair{x}{\lpair{y}{\lpair{u}{v}}} : \leftapp{\rightapp{y}{\rightapp{\leftapp{u}{x}}{v}}}{x}$
\end{tabular}
}
\caption{Derivation of \quotes{x loves Gary his code} (left) and \quotes{x loves his code} (right)}
\label{fig:subproofs}
\end{figure}
\end{center}
%\vspace*{\fill}
%\end{landscape}
%\newpage
%\begin{landscape}
%\topskip0pt
%\vspace*{\fill}
\begin{center}
\begin{figure}[h]
\resizebox{\linewidth}{!}{
  \def\FORMULA{\formulaone}%
  \def\SUBPROOF{%
  \def\FORMULA{(np_{} \bs s_{}) \s np_{} \tensor (\Diamond np_{} \bs (np_{} \s n_{}) \tensor n_{})}
  \def\SUBPROOFONE{\infer*[]{\overset{Gary}{np} \tensor (\overset{loves}{(np \bs s) \s np} \tensor (\overset{his}{\fdia np \bs (np \s n)} \tensor \overset{code}{n})) \longrightarrow s}{}}%
  \def\SUBPROOFTWO{%
  \def\FORMULA{\formulaone}%
  \def\SUBPROOF{\infer[\fdia]{\fdia (loves(np \bs s) \s np \tensor (\fdia np \bs (np \s n) \tensor n)) \longrightarrow \fdia (np \bs s)}{\infer[\resleft]{(np \bs s) \s np \tensor (\fdia np \bs (np \s n) \tensor n) \longrightarrow np \bs s}{\infer*[]{\overset{Bill}{np} \tensor (\overset{loves}{(np \bs s) \s np} \tensor (\overset{his}{\fdia np \bs (np \s n)} \tensor \overset{code}{n})) \longrightarrow s}{}}}}%
  \input{derivations/itv_vp_site_param}%
}%
  \input{derivations/coordinate_phase_param}%
}%
  \input{derivations/struct_phase_param}%

}
\caption{Sloppy reading of gary's code. \quotes{loves his code} is copied to the two main subproofs, where each one is resolved with their respective noun phrase argument (left: Gary, right: Bill).}
\label{fig:sloppy}
\end{figure}
\end{center}
%\vspace*{\fill}
%\end{landscape}
%\newpage
%\begin{landscape}
\begin{center}
\begin{figure}[h]
\resizebox{\linewidth}{!}{
  \def\FORMULA{\formulathree}%
  \def\SUBPROOF{%
  \def\FORMULA{\formulatwo}%
  \def\SUBPROOF{%
  \def\FORMULA{\formulatwo}
  \def\SUBPROOFONE{\infer*[]{\overset{Gary}{np} \tensor (\overset{loves}{(np \bs s) \s np} \tensor ((\overset{\redbox{Gary}}{\fdia np} \tensor \overset{his}{\fdia np \bs (np / n)}) \tensor \overset{code}{n})) \longrightarrow s}{}}%
  \def\SUBPROOFTWO{%
  \def\FORMULA{\formulatwo}%
  \def\SUBPROOF{\infer[\fdia]{\fdia ((np \bs s) \s np \tensor ((\fdia np \tensor \fdia np \bs (np / n)) \tensor n)) \longrightarrow \fdia (np \bs s)}{\infer[\resleft]{(np \bs s) \s np \tensor ((\fdia np \tensor \fdia np \bs (np / n)) \tensor n) \longrightarrow np \bs s}{\infer*[]{\overset{Bill}{np} \tensor (\overset{loves}{(np \bs s) \s np} \tensor ((\overset{\redbox{Gary}}{\fdia np} \tensor \overset{his}{\fdia np \bs (np / n)}) \tensor \overset{code}{n})) \longrightarrow s}{}}}}%
  \input{derivations/itv_vp_site_param}%
}%
  \input{derivations/coordinate_phase_param}%
}%
  \input{derivations/struct_phase_param}%
}%
  \input{derivations/strict_final_phase}%

}
\caption{Strict reading of gary's code. First, \quotes{Gary} is copied and resolved with \quotes{loves his code}, after which \quotes{loves Gary his code} is copied to the two main subproofs (left: Gary, right: Gary).}
\label{fig:strict}
\end{figure}
\end{center}
\end{landscape}

\end{document}